\documentclass[lettersize,journal]{IEEEtran}
\usepackage{amsmath,amsfonts}
\usepackage{algorithmic}
\usepackage{algorithm}
\usepackage{array}
\usepackage[caption=false,font=normalsize,labelfont=sf,textfont=sf]{subfig}
\usepackage{textcomp}
\usepackage{stfloats}
\usepackage{url}
\usepackage{verbatim}
\usepackage{graphicx}
\usepackage{cite}
\usepackage{caption}
\usepackage{xcolor}

\ifodd 0
\definecolor{darkgreen}{RGB}{0,200,0}
\newcommand{\rev}[1]{{\color{blue}#1}} %revise of the text
\else
\newcommand{\rev}[1]{#1}
\fi

\begin{document}

\title{Split Learning in 6G Edge Networks}
\author{Zheng Lin, Guanqiao Qu,  Xianhao Chen,~\IEEEmembership{Member,~IEEE}, and Kaibin Huang,~\IEEEmembership{Fellow,~IEEE}
%\thanks{Guangyu Zhu is with the Department of Electrical and Computer Engineering, University of Florida, Gainesville, FL 32611 USA (e-mail: gzhu@ufl.edu).}
%\thanks{Yiqin Deng is with the School of Control Science and Engineering, Shandong University, Jinan 250061, Shandong, China (e-mail: yiqin.deng@email.sdu.edu.cn).}

\thanks{Zheng Lin, Guanqiao Qu, Xianhao Chen, and Kaibin Huang are with the Department of Electrical and Electronic Engineering, University of Hong Kong, Pok Fu Lam, Hong Kong (e-mail: linzheng@eee.hku.hk; gqqu@eee.hku.hk; xchen@eee.hku.hk; huangkb@eee.hku.hk).
\textit{(Corresponding author: Xianhao Chen)}}}
%
%\thanks{Yue Gao is with the School of Computer Science, Fudan University, Shanghai 200438, China (email: gao\underline{~}yue@fudan.edu.cn).}
%
%\thanks{Yuguang Fang is with the Department of Computer Science, City University of Hong Kong, Kowloon, Hong Kong (e-mail: my.fang@cityu.edu.hk).

\markboth{}%
{Shell \MakeLowercase{\textit{et al.}}: A Sample Article Using IEEEtran.cls for IEEE Journals}


\maketitle

\begin{abstract}
With the proliferation of distributed edge computing resources, the 6G mobile network will evolve into a network for connected intelligence. Along this line, the proposal to incorporate federated learning into the mobile edge has gained considerable interest in recent years. However, the deployment of federated learning faces substantial challenges as massive resource-limited IoT devices can hardly support on-device model training. This leads to the emergence of split learning (SL) which enables servers to handle the major training workload while still enhancing data privacy. In this article, we offer a brief overview of SL and articulate its seamless integration with wireless edge networks. We begin by illustrating the tailored 6G architecture to support split edge learning (SEL). Then, we examine the critical design issues for SEL, including resource-efficient learning frameworks and resource management strategies under a single edge server. Furthermore, from a networking perspective, we expand the scope to multi-edge scenarios, exploring multi-edge collaboration and model placement/migration. Finally, we discuss open problems for SEL, including convergence analysis, asynchronous SL, and label privacy preservation.
\end{abstract}
\vspace{-0.1cm}
\vspace{-0.1cm}
\section{Introduction\label{introduction}}
\vspace{-0.1cm}
The conventional cloud-based model training, which centralizes all data for processing, is no longer sufficient to meet the soaring data traffic demands, ubiquitous computing needs, stringent latency, and personalization requirements of emerging Internet-of-Things (IoT) applications. To overcome these challenges, edge learning has emerged as an exciting research direction, which harnesses the power of multi-access edge computing (MEC) to support machine learning and local training, thereby achieving reduced backhaul bandwidth costs, ultra-low latency, and context awareness. For example, as a subdomain of edge learning, federated edge learning (FEEL) has attracted significant research and industry interest in the past few years due to its privacy-enhancement nature, which has already been discussed in 3GPP release 18 for 5G standardization~\cite{Release18}. 

%breaking down the "data silos" that arise due to data access regulations (e.g., GDPR and CCPA) and data owners' privacy concerns. 

\begin{figure*}[t!]
\centering
\includegraphics[width=14.3cm]{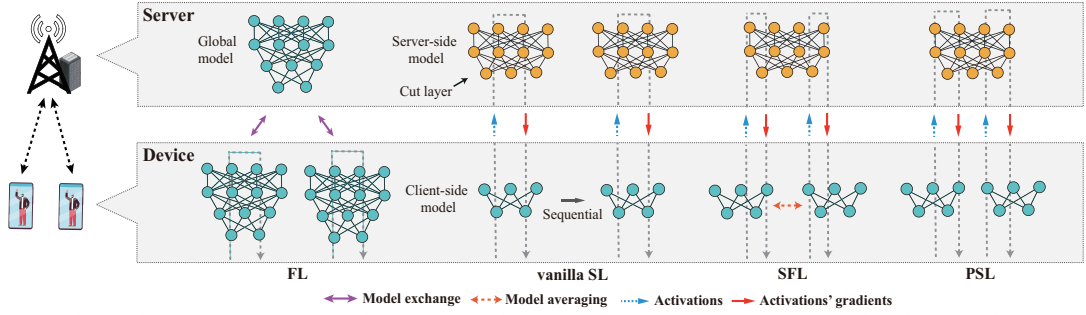}
\vspace{-0.2cm}
\caption{\rev{An illustration of FL and the state-of-the-art SL frameworks, where model averaging in SFL is conducted on a fed server located close to clients~\cite{thapa2022splitfed}.}
}
\label{DCML_comparison}
\end{figure*}

However, federated learning (FL) also has its limitations. The core idea of FL is to leverage local model training and global model aggregation for collaborative learning without accessing users' raw data, as illustrated in Fig. \ref{DCML_comparison}. Unfortunately, FL may be infeasible for a large number of resource-limited IoT devices since the entire model is trained at end devices. To overcome this hurdle, split learning (SL) has emerged as a promising model training scheme. By splitting a model and placing a part of it at an edge server, SL allows a server to handle the major workload of deep neural networks (DNNs) based on model splitting while still retaining a few early layers and raw data at local devices for privacy preservation. This approach significantly reduces the computing, storage, and memory requirements for model training, making machine learning (ML) more accessible to resource-constrained devices.

%communication overhead because the size of the layer width, in many cases, could be smaller than the raw data, such as High-Definition (HD) images, thereby reducing bandwidth usage and latency across communication networks.
%In contrast to 5G, which mainly supports basic AI functionalities, 

As an alternative/complementary approach to FL, we believe that SL has the potential to become one of the predominant AI technologies in the 6G edge. 6G will be a network of sensors, computing devices, and ML to achieve superior performance for ubiquitous AI tasks while addressing concerns of data ownership and privacy~\cite{huawei20216g}. The favorable characteristics of SL are perfectly aligned with the vision of 6G. On the one hand, as explained earlier, \textit{SL allows for training workload offloading while enhancing data privacy}. This is of paramount importance for pervasive mobile and IoT devices with constrained hardware, such as mobile phones and smart cameras, which may struggle to support compute-intensive local training as required in FL. It is also difficult to recover the raw data in SL due to the exchange of activations and the server's lack of knowledge of client-side models. On the other hand, \textit{SL enables better resource utilization by leveraging dispersed computing and memory resources over the network edge}.  Nowadays, it is common for deep neural networks to contain millions or even billions of parameters, making them challenging to train at the edge. For instance, the large language model (LLM) 7B LLaMA, which is the smallest version of LLaMA feasible for on-device deployment, still has 7 billion parameters. In the 6G era, edge computing resources will become ubiquitous, where servers can be the ones in the network core, macro base stations, small base stations, pico base stations, and even autonomous cars and mobile phones. By enabling model splitting, SL at the 6G edge can facilitate flexible computing load sharing among end devices and multiple servers for collaborative learning. This enables the best utilization of distributed computing resources for performing resource-intensive edge training.

The integration of mobile edge networks and SL presents unique technical challenges and exciting research opportunities. To date, this field remains relatively under-explored. To facilitate effective SL at the edge, the main design challenges arise from the need for frequent transmissions of high-dimension features and back-propagated gradients over bandwidth-limited wireless channels and training deep models over resource-constrained edge networks. These problems can only be properly addressed by a holistic design of SL and communication-computing resource management, which is highly challenging. To our best knowledge, this is the first article that focuses on how to effectively support split edge learning (SEL) in resource-constrained wireless networks.

This article aims to thoroughly examine the deployment of SL in mobile edge networks. To this end, we first envision the 6G architectural design tailored for SL. Second, we identify the potential directions for efficient SL design that leverages model compression, activation compression, and back-propagated to decrease the resource consumption of SL. Third, we introduce innovative designs for resource management issues related to SEL under both single-edge and multi-edge scenarios, such as dynamic resource allocation to optimize idle resource utilization, hierarchical or multi-hop SL for collaborative training of large models, and model placement/migration strategies to accommodate user distribution and mobility. In summary, this article provides the first comprehensive review of SEL and highlights research opportunities. It is important to note that despite the similarity in model splitting, the problems for SEL significantly differ from edge split inference which has been covered by some existing articles~\cite{shao2020communication}, as the goal of SEL is model training/fine-tuning rather than predictions at the edge.

\rev{This article is organized as follows. We begin by examining the existing SL approaches in Section \ref{background}. Our discussion will then shift to the synergy between SL and 6G edge. Particularly, we elaborate on the 6G architectural design for SL in Section \ref{architecture}, followed by in-depth discussions on innovative resource-efficient SL framework design in Section \ref{efficientSL}. We present the resource management strategies for SL under single-cell and multi-edge scenarios in Section \ref{single} and Section \ref{multi}, respectively. Finally, we identify open problems for SEL in Section \ref{Open} and conclude this article in Section \ref{conclusion}.}

% We present the resource management strategies for SL under single-cell scenarios in \ref{single}. Moreover, from a networking perspective, we identify the design for SL under multi-edge scenarios in \ref{multi}. Finally, we identify open problems for SEL in Section \ref{Open} and conclude this article in Section \ref{conclusion}.
\vspace{-0.2cm}
\section{Background\label{background}}
\vspace{-0.1cm}
\rev{In this section, we briefly introduce several existing SL approaches and compare them with FL.}

% In this section, we briefly introduce several existing SL approaches and compare them with FL.
\vspace{-0.2cm}
\subsection{Federated learning}
\vspace{-0.1cm}
Standard machine learning approaches generally run optimization algorithms like Stochastic Gradient Descent (SGD) in a remote cloud center with centralized training data, resulting in severe privacy leakage. To mitigate this issue, Google proposed FL in 2016, enabling mobile phones to collaboratively learn a shared model while keeping all the training data locally~\cite{mcmahan2017communication}. Specifically, devices only train and upload the model updates to the server for aggregation and download the aggregated model from the server, as shown in Fig. \ref{DCML_comparison}. 
Due to the huge size of models and the repeated model upload/download, FL suffers from significant communication latency and exerts a tremendous burden on telecommunication infrastructure. For this reason, FL has sparked considerable interest from the telecommunication industry, which aims to implement communication-efficient FL deployment at the mobile edge, leading to the emergence of a new field known as "federated edge learning" (FEEL).

\vspace{-0.2cm}
\subsection{Split learning}
\vspace{-0.1cm}
Given the increasing size of models and resource-limited edge devices, FL may not be suitable for various intelligent applications due to the need for full model training on devices. As introduced in 2018, SL has emerged as a privacy-enhancing collaborative learning framework that addresses resource limitations while preserving data privacy. The idea is simply to partition a model into two or more parts and place them on the client and server sides, respectively, enabling the server to share the training workload.

There are several variants of SL. Vanilla SL, which is the original form, operates sequentially, which trains the model for one client at a time~\cite{vepakomma2018split}. However, the sequential training process of vanilla SL incurs excessive training latency. Moreover, under highly non-IID settings, the sequential training could yield poor learning performance as the model tends to better fit the data distribution of the last client.

To address these issues, split federated learning (SFL)~\cite{thapa2022splitfed} and parallel split learning (PSL)~\cite{kim2022bargaining,joshi2021splitfed} have been devised to parallelize client-side model training, empowering clients to train their sub-models simultaneously, as illustrated in Fig.~\ref{DCML_comparison}. The distinction between SFL and PSL lies in the synchronization requirements: following the spirit of FL, the former requires model averaging for client-side models periodically whereas the latter does not require it. As a result, SFL results in increased communication costs due to client-side model transfer. Moreover, SFL requires separate and non-colluding servers for server-side model training and client-side model averaging, respectively. Otherwise, with the output of the client-side model and the client-side model parameters, an adversary server can easily recover the input raw data. Conversely, PSL eliminates the need for client-side model synchronization, which overcomes the above limitations. Yet, it naturally results in varied client-side model parameters across devices, which may adversely impact training convergence in comparison to SFL, as shown in \cite{lin2023efficient}.

\begin{figure*}[t!]
\centering
\includegraphics[width=14.5 cm]{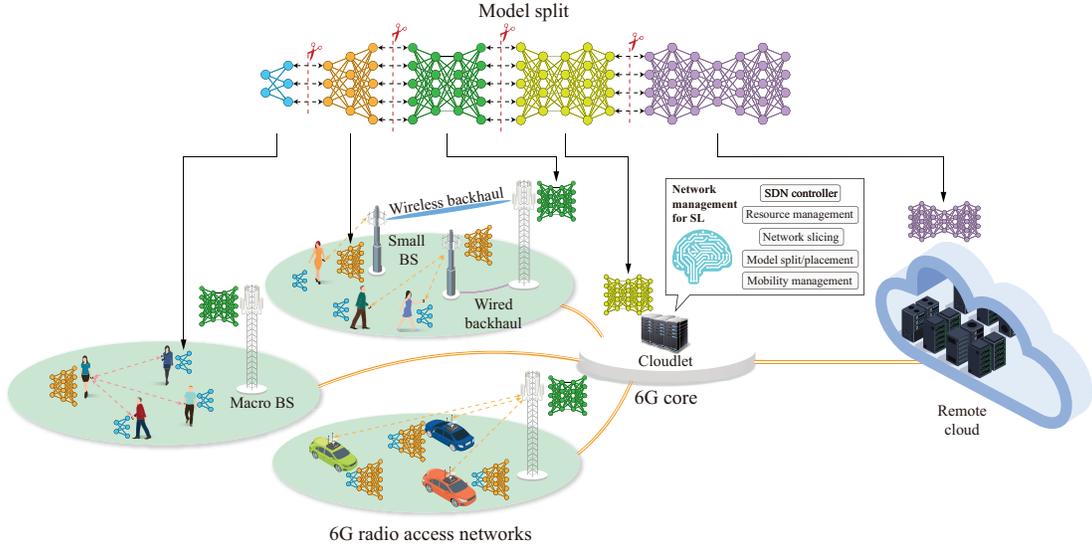}
\vspace{-0.1cm}
\caption{The overall architecture for SL in 6G.
}
\label{hierarchical_SL}
\end{figure*}

\vspace{-0.2cm}
\subsection{Federated Learning v.s. Split Learning} 
\vspace{-0.1cm}
A natural question is how to choose between SL and FL, both of which are privacy-enhancing collaborative learning frameworks. A crucial factor is the computing capability of end devices. As previously mentioned, SL is a natural solution when the end device is resource-limited, which can hardly train a large model due to computing and memory constraints~\cite{lin2023pushing}.

From a communication perspective, when the training dataset from clients is large, FL may be preferable as SL incurs large volumes of smashed data that is proportional to the size of the dataset. \rev{In contrast, since FL involves model transmissions, SL becomes more communication-efficient when the data size of the model is larger than that of smashed data (e.g., for ResNet-152, when batch size is 32, the 37-th layer smashed data volume is approximately 0.49 MB, the model size is however 230 MB.). 
The detailed comparison of FL and the state-of-the-art SL frameworks is illustrated in Table 1 in~\cite{lin2023efficient}.}

\section{Architectural Design for Split  Edge Learning\label{architecture}}
SEL demands a holistic design of communications and training because there exists a fundamental tradeoff between the computing cost for training sub-models and the communication cost for transmitting smashed data (i.e., intermediate activations/back-propagated gradients) between the collaborating devices. A well-designed supporting architecture is essential to optimize training convergence under a resource-constrained wireless network. 6G, which builds upon 5G, features the true convergence of communications and computing, providing opportunities to advance the integration of MEC and SL. \rev{With this in mind, we will envision the potential 6G network architecture tailored for SL.}

Due to resource limitations, a single computing device/server may not be capable of training/deploying a large AI model. To enable effective SL service provisioning, a hierarchical system is necessary to handle services with different computing loads, delay constraints, and personalization requirements. 6G SL system comprises data sources, such as smart cameras, mobile phones, and autonomous cars, and heterogeneous cellular base stations, such as small and macro base stations, edge servers (cloudlets) located at cell aggregation sites, and the remote cloud. As a result, versatile AI models, including complete or partial models, are distributed across multiple levels of the system for collaborative training. In general, a larger (sub) model can be placed at a more powerful node further away from data sources whereas a smaller (sub) model can be stored at the resource-constrained edge devices/servers. The high-level computing nodes can also store (sub) models with more general representations for usage by users at a large scale, while lower-level edge nodes store (sub) models better fitting local environments. Depending on the configurations, smashed data can be exchanged between these devices/servers through device-to-device (D2D), vehicle-to-vehicle (V2V) links, wireless backhaul, or wired backhaul, as illustrated in
the Fig. \ref{hierarchical_SL}. 

%Depending on the configurations, models or smashed data can be exchanged between these devices/servers through device-to-device (D2D) or vehicle-to-vehicle (V2V) links, wireless backhaul, or wired backhaul for collaborative learning/inference, as illustrated in the figure. 

In SL, model training across multiple computing devices/servers requires judicious resource coordination. To provide end-to-end QoS guarantees, centralized control is arguably indispensable. Fortunately, the central intelligence is aligned well with 6G architecture. To achieve this, the 6G edge can implement software-defined networking (SDN) to facilitate model transfer, smashed data routing, and computing resource allocation. By monitoring network link status, and computing/storage/memory resource availability, the central controller proactively splits models, manages computing and networking resources, configures data routing, and conducts model placement/migration. End-to-end network slicing can be utilized for various SL tasks to achieve differentiated QoS provisioning. For example, SL tasks for autonomous driving or robot control demand ultra-low latency while SL tasks for training a next-word prediction model may not be time-sensitive. Finally, the 6G edge should also have mobility management components to enable seamless service and model migration, allowing context-aware models to follow users as users move. By implementing these approaches, the 6G system can leverage network-wide distributed resources and meet the QoS requirements of diverse SL applications.
\vspace{-0.1cm}
\section{Resource-efficient Split Learning Frameworks\label{efficientSL}}
\vspace{-0.1cm}
Despite the promising benefits, the limited spectrum and computing resources at the network edge pose significant hurdles to the effective implementation of SL. It is often worthwhile to trade off training accuracy for reduced latency under limited networking and computing resources. \rev{In what follows, we present innovative SL frameworks that decrease the resource demands in different aspects.}

% In what follows, we present innovative SL frameworks to decrease the resource demands of SL in different aspects.

%This section focuses on the most common scenario: SL under a single cell or a single edge server, and delves into two-fold solutions: 1) develop more resource-efficient training frameworks, and 2) manage the network resources more effectively. We will also discuss other variants of SL under single-edge systems.

%In what follows, we illustrate resource-efficient SL from three aspects, aiming to reduce server-side training workload, communication costs, and device-side training workload, respectively.
\vspace{-0.1cm}
\subsection{Split Learning with Activation Compression}
\vspace{-0.1cm}
First of all, it is of paramount importance to reduce communication overhead for smashed data exchange over the split layer between devices and the edge server. To mitigate this issue, one promising direction is to adopt an auto-encoder, which trains an encoder to compress the data and then a decoder to recover the data~\cite{hsieh2022c3}. In \cite{hsieh2022c3}, cyclic convolution is employed to compress multiple features into a single compressed feature, which is
decoded on the server side through cyclic correlation. Although the process introduces noise, the impact on learning performance is shown to be small.

As neural networks, auto-encoders bring additional computation and training costs and are challenging to understand theoretically. Thus, the other direction is to directly compress the smashed data. To this end, feature compression has been explored in split inference by pruning activations~\cite{shao2020communication}. Nevertheless, its impact on SL is worth further exploration. Also, it is important to theoretically characterize the convergence bound of SL in terms of compression ratios, based on which an SL scheme with a carefully designed feature compression ratio can be developed to achieve the optimal balance between training accuracy and latency.

%This presents two conflicting goals: reducing data upload latency by compressing smashed data, and accelerating training using high-precision data. Since the compression of the activations results in zero gradients during the backpropagation due to the rounding operation in a piece-wise flat operator, one promising direction is to use the straight-through estimator to approximate the gradients~\cite{yin2019understanding} to overcome the non-differentiable quantization operator. 

\vspace{-0.2cm}
\subsection{Split Learning with Weight Compression} 
\vspace{-0.1cm}
The second problem is how to reduce the computing workload, especially on devices. Even though a server handles the majority of the workload in SL, the remaining computing load, such as several early layers necessary for concealing the raw data, might still be too demanding for resource-constrained mobile/IoT devices. A feasible solution is training on compressed models to further reduce computing and memory costs. There are several popular compression techniques\cite{deng2020model}: \rev{ \romannumeral1) \textit{Model quantization} reduces the bitwidths of both weights and activations (e.g., from full precisions to 8 bits) to lower the training latency and memory requirements;} \romannumeral2) \textit{model pruning} directly eliminates the number of parameters of DNN. Our goal is to decrease computation costs by directly training on compressed weights as well as reduce communication overhead by reducing the bitwidths of activations at the cut layer. In SL, it is beneficial to devise a model compression scheme that allows clients and the edge server to use varied compression ratios based on their computing capabilities. For example, since a server can be much more powerful than a client, clients can train on a low-precision sub-model, while a server can handle a high-precision sub-model. Theoretically analyzing the impact of this scheme is an interesting topic for future research, which can offer guidance on how to deploy SL with compression in resource-constrained settings.

%\romannumeral3) \textit{low-rank factorization} (e.g., singular value decomposition (SVD)) replaces the neural network weight matrix with matrices of a smaller dimension. 

%There are several popular compression techniques. \romannumeral1) \textit{Model quantization} reduces the number of bits used to represent both weights and activations (e.g., from full precisions to 8 bits) to reduce the number of MAC operations and the memory requirements. \romannumeral2) \textit{model pruning} directly eliminates the number of parameters of DNN. \romannumeral3) \textit{low-rank factorization} (e.g., singular value decomposition (SVD)) replaces the neural network weight matrix with matrices of a smaller dimension. 
%However, the decomposition operation might be compute-expensive. 

\begin{figure}[t]
\centering
\includegraphics[width=0.37\textwidth]{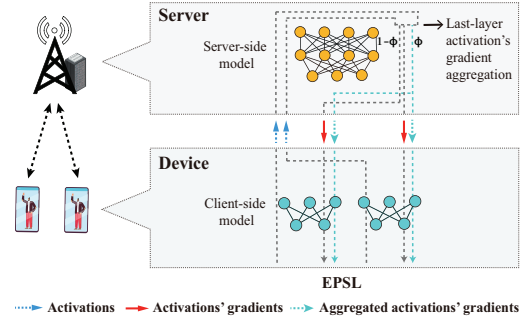}
\vspace{-0cm}
\caption{The proposed EPSL scheme in \cite{lin2023efficient}.
}
\label{EPSL_scheme}
\end{figure}

\begin{figure}[t]
\centering
\includegraphics[width=5 cm]{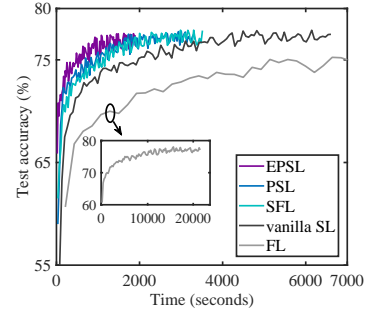}
\vspace{-0.2cm}
\caption{The test accuracy of ResNet-18 on HAM10000 dataset under FL, vanilla SL, SFL, PSL and EPSL. The data samples are distributed over $5$ clients under IID settings, where the total available bandwidth is 70 MHz, the computing capability of each client is uniformly distributed within $[0.1, 0.5] \times 10^9$ cycles/s, and the computing capability of the server is $7 \times 10^9$ cycles/s.
}
\label{EPSL_training_performance}
\end{figure}
\subsection{Split Learning with Back-propogated Gradient Aggregation}
\vspace{-0.1cm}
The final challenge we need to address is reducing server-side computing workload. Although an edge server is generally more powerful than an edge device, it can also become the bottleneck in PSL/SFL since the server may serve a massive number of clients and often take over the majority of the training workload. To tackle this issue, we have proposed efficient parallel split learning (EPSL)~\cite{lin2023efficient} to reduce the dimension of back-propagated gradients by aggregating them at the last layer, as depicted in Fig. \ref{EPSL_scheme}. Compared with existing state-of-the-art SL benchmarks, such as SFL and PSL, this method can reduce back-propagation computing and communication costs from $\mathcal O(M)$ (number of clients) to $\mathcal O(1)$. Note that EPSL can also control the aggregation ratio $\phi$ in the backpropagation process to strike a balance between the reduction in communications/computing costs and learning accuracy, where $\phi = 0$ reduces EPSL to PSL. The superiority of EPSL over other SL approaches is demonstrated in Fig. \ref{EPSL_training_performance}, where the back-propagated gradients are reduced without noticeably impacting the learning accuracy (i.e., with 0.46$\%$ deterioration when the model converges). More details can be found in \cite{lin2023efficient}.

\section{Resource Management for Split Learning: The Single-cell Perspective\label{single}}
\vspace{-0.1cm}
\begin{figure}[!t]
\centering
\includegraphics[width=0.44\textwidth]{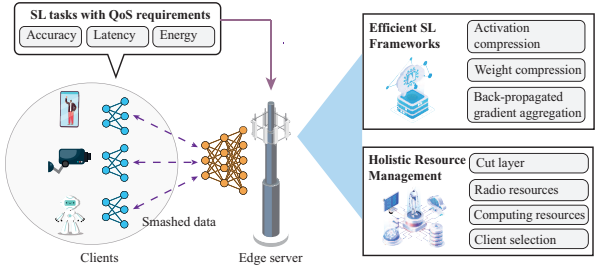}
\caption{An illustration of SL under single-cell systems.
}
\label{single_server}
\end{figure}
In parallel split learning, the training latency is determined by the slowest client, also known as the "straggler." To mitigate this issue, the channels and server-side computing resources should be judiciously allocated to the stragglers to optimize the training process. Although the straggler effect is also present in FL, parallel split learning involves model splitting and smashed data exchange, making the design significantly different from the approaches for FL. \rev{In light of these needs, we will discuss network resource allocation problems tailored for SEL under a single cell as shown in Fig. \ref{single_server}.}

\vspace{-0.2cm}
\subsection{Joint Resource Allocation and Model Split} 
\rev{Network resource allocation is tightly coupled with model splitting in SL, distinguishing it from FL.} The split layer significantly impacts training latency, which can result in varied training workloads between devices and edge servers and different communication overheads due to the size of layer output. In particular, when splitting the model at a "deeper" layer, more computing workload is left on the client side, while the communication overhead can be potentially reduced as the size of layer output often shrinks as it traverses deeper, such as in most convolutional neural networks (CNNs). \rev{Consequently, the joint optimization of model splitting and resource allocation is essential to strike a good balance between computing and communication resources.} On the other hand, since an edge server in SL supports parallel training for multiple clients, allocating more computing and channel resources to the straggler is necessary to compensate for its limited local computing and communication capabilities. Along this line, Wu et al. propose a cluster-based SL in which clients concurrently train the model in each cluster based on SFL~\cite{wu2023split}. Subsequently, the model undergoes training across different groups based on the traditional SL method. This approach stochastically optimizes the cut layer selection, device clustering, and radio spectrum allocation, where the cut layer selection decision is made in a larger timescale whereas device clustering and radio spectrum allocation decisions are made in a smaller timescale. Taking a step further, it is vital to develop on-demand resource scheduling schemes for PSL/SFL. The existing solutions allocate fixed resources for each client during a training round~\cite{lin2023efficient,wu2023split}. However, this static resource partitioning leaves resources idle for a significant portion of the time in SL. For instance, when a client performs forward propagation, there is no data to transmit, leaving the assigned channels and server-side computing resources idle. \rev{Unlike FL with a fixed data size for exchanged data (i.e., the size of the model), SL has the flexibility to control the computing and communication overhead via batching.} Therefore, exploring on-demand resource scheduling for PSL/SFL, which \textit{dynamically} allocates channel and computing resources to clients in need to minimize latency, is worth further investigation. 

%Furthermore, since local training time greatly depends on batch sizes, we can optimize the batch size of each client so that they can access wireless channels at different time instants, thereby uploading data with high speed.
%This inefficiency is analogous to circuit-switched networks that do not utilize network resources efficiently compared to packet-switched networks because the former allocates resources to a session even though there is no data to transmit.
\vspace{-0.2cm}
\subsection{Client Selection}
\vspace{-0.1cm}
Due to resource limitations, selecting all active clients for training may be impractical. Considering partial client participation, client selection plays a crucial role in SEL. The 6G edge demands a unified client selection framework taking two factors into account: 1) the number of selected clients (or training data samples) and 2) the data diversity. On the one hand, some works for distributed learning aim to select as many as clients of resource heterogeneity as possible under deadline requirements~\cite{chen2022federated}. The rationale is that involving more participants (or equivalently, more data samples) joining the training generally leads to faster convergence speed. On the other hand, maximizing the number of clients can result in a biased model, because in this case, client devices with poor channel conditions (e.g., at the cell edge) and limited computing capabilities are likely to be excluded. Therefore, it is also necessary to select clients based on their data distributions. \rev{Unlike FL, SL can select clients based on smashed data, which is essentially high-level features of original data. A promising idea is to select a set of clients with smashed data better representing the global smashed data distributions. The effectiveness of this strategy demands further validation.} Note that ensuring data diversity could contradict the goal of selecting more clients within a deadline. The unified client selection framework is expected to balance the number of clients selected and the data diversity for SL.

\vspace{-0.1cm}
\section{Resource Management for Split Learning: The Networking Perspective\label{multi}}
\vspace{-0.1cm}
The growing size of AI models presents a substantial challenge for edge learning. Based on multi-edge split learning, we can deploy large models at the 6G edge while overcoming the computing and memory constraints through sharing the workloads among distributed edge servers. Furthermore, model placement and migration are anticipated to be basic components of SEL. This section is devoted to these aspects, which examine SEL from a networking perspective. 

\vspace{-0.3cm}
\subsection{Hierarchical Split Learning}
\begin{figure}[t!]
\centering
\includegraphics[width=5 cm]{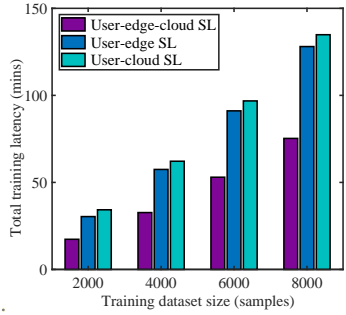}
\vspace{-0cm}
\caption{The total training latency of user-edge-cloud, user-edge, and user-cloud architectures for achieving target accuracy on HAM10000 versus the training dataset size. The data samples are distributed over $5$ clients under IID settings,  where the computing capability of the cloud is $20 \times 10^9$ cycles/s, the edge-cloud link capacity is set to $\frac{1}{20}$ of the user-edge link capacity~\cite{9562523}, and other key parameters are consistent with Fig.~\ref{EPSL_training_performance}.
}
\label{different_datasize}
\end{figure}

\vspace{-0.1cm}
The practical 6G systems feature hierarchical computing architecture with cloud/edge servers of various levels, as illustrated in Fig. \ref{hierarchical_SL}. To facilitate effective learning, it is crucial to coordinate multi-tier resources. It is important to note that, in comparison to the more common two-level SL, multi-level server collaboration provides greater flexibility in achieving a balanced trade-off between communication and computing. To demonstrate the effectiveness of multi-level SL, let us consider a three-tiered user-edge-cloud architecture. In this case, the communication bottleneck and latency often lie in the edge-cloud link. Meanwhile, as noted earlier, the layer size tends to diminish as it progresses deeper in many practical models like CNNs. Based on this observation, assigning some layers to end users and some other layers to the edge server allows for a deep and more ``narrow'' split layer between the edge and cloud, thereby reducing communication costs. In contrast, a two-tiered user-cloud architecture could involve excessive communication latency due to a large volume of smashed data exchange with the cloud for a ``wider'' early split layer (as the user can only execute several early layers). On the other hand, the two-tiered user-edge architecture, in spite of eliminating the need for cloud-edge transmissions, lacks adequate computing power at the edge. By considering PSL with five clients, Fig. \ref{different_datasize} demonstrates the superiority of the hierarchical cloud-edge-user SL architecture over these two-tiered counterparts. To this end, exploring hierarchical SL with potentially more levels for large-scale users is a promising research direction.

%Appropriate resource management (channel and computing resource management) facilitates better adaption of SL to wireless edge systems, which bolsters learning performance (e.g., speeds up model training or improves energy efficiency). 
%In addition to partitioning the training workload between end devices and edge servers, the cut layer selection also results in various communication overheads (stemming from different output data sizes for each layer). Thus, joint model split and resource scheduling strategies merit further investigated in SEL.
\vspace{-0.2cm}
\subsection{Multi-hop Split Learning}
\vspace{-0.1cm}
We extend our considered scenarios to the general mesh network. The aforementioned hierarchical SL is a type of multi-hop SL, yet confined to the "vertical" paradigm only consisting of servers of different levels. In a more general sense, numerous small/macro base stations can form a mesh of edge servers for multi-hop split learning. The primary motivation is to better share the workload among multiple servers to handle compute-intensive model training. To optimize the performance, it is essential to examine the joint system design of \textit{model splitting and data routing} in multi-hop edge computing networks, taking into account bandwidth, computing, and memory constraints. In the 6G mobile networks, centralized smashed data routing can be implemented by considering sub-model splitting/placement and computing/bandwidth resource constraints in a centralized manner with SDN. This approach is expected to be more effective than distributed routing due to the global knowledge obtained by the central controller.

In addition to static edge servers, multi-hop SL can also be implemented in mobile ad hoc networks within the 6G paradigm. For example, a vehicular platoon can implement SL by partitioning and sharing a model among the vehicles within the group based on vehicle-to-vehicle (V2V) communications. Based on device-to-device (D2D) communications, smartphone users can also train a large model by splitting it into several parts. All these scenarios could capitalize on the dispersed resources at the network edge, as a single device/edge server may not be able to handle compute-intensive training tasks individually.

% The tremendous volume of data generated on client devices will give rise to a massive cut-layer data exchange between client devices and servers, posing a formidable impediment to SL deployment at the wireless edge. The majority of the existing research in this domain concentrates on single-hop wireless networks, where multiple client devices collaborate with an edge server for model training.  Nevertheless, owing to the limitations imposed by transmit power, clients located far from the edge server may encounter difficulties in data exchange or suffer from diminished uplink data rates due to severe path loss. The wireless multi-hop networks offer a cost-effective deployment alternative for seamless service provision across expansive areas.

%First, judicious spectrum allocation is essential to mitigate network-wide wireless interference. Second, cut-layer data routing must be optimized by considering heterogeneous computing and communication capabilities. Finally, the cut layer selection emerges as a critical factor, entwined with computational workload offloading and the volume of data exchanged. Therefore, joint spectrum allocation, routing, and model partitioning optimization becomes indispensable for facilitating SL deployment in multi-hop networks.

%\subsubsection{Decentralized Split Learning}
\vspace{-0.2cm}
\subsection{Edge Model Placement and Migration}
\vspace{-0.1cm}
The 6G network edge processes distributed storage resources, which can be explored for the placement and migration of versatile AI models to facilitate SL operations. Split learning/inference can leverage ``partial model placement'' to enhance the caching performance due to the fact that users and servers can execute part of the neural networks. Therefore, there exists a tradeoff between edge storage and communication and computing costs. While placing a larger portion of a model at the edge node occupies more storage space, it potentially reduces communication costs for exchanging data with other nodes. \rev{Therefore, it is crucial to jointly design model splitting and model placement for service placement/migration, considering bandwidth, computing and memory constraints.} Besides, model placement/migration in SL can account for time-varying geographical data distributions of clients. There exist general (partial) models that suit a broad range of users/services (e.g., autonomous driving) but lack supreme task-specific performance, and also fine-tuned (partial) models specialized at certain tasks (e.g., autonomous driving for urban environments under rainy days). In model training/inference, the appropriate placement of these (partial) models enables real-time and low-cost data/model transfer between data sources and computing servers. These factors necessitate a revisit of service placement/migration problems under the edge computing paradigm.

%The concept of content/service placement caching has been widely explored in the literature, where the objectives are to optimally place popular content, algorithms, or data libraries at edge servers to facilitate content downloading or computing service provisioning. Similarly, in split learning/inference, versatile AI models should be proactively cached at edge nodes to accommodate potential users' requests. In split learning/inference, versatile AI models should be proactively cached at edge nodes to accommodate potential users' requests. 

%since users will have more freedom to choose which layer to split or even transmit the raw data. 
\vspace{-0.2cm}
\section{Open problems\label{Open}}
\vspace{-0.1cm}
Although we have highlighted some research challenges and solutions, there are still a few pressing research issues. We discuss these open problems as follows.

\vspace{-0.2cm}
\subsection{Convergence Analysis for Parallel Split Learning}
\vspace{-0.1cm}
Convergence analysis plays a pivotal role in resource optimization for SEL, as it guides us to allocate resources to accelerate training. Essentially, PSL can be regarded as a special case of SFL where client-side models will never be aggregated. In general, the convergence of SFL still requires further understanding, especially on how client-side model aggregation will impact model convergence. In the extreme case, PSL eliminates the need for client-side model aggregation, resulting in the same server-side model and varied client-side models across devices. Although empirical experiments have demonstrated that its impairment to learning performance appears to be small~\cite{lin2023efficient}, to our best knowledge, there is no theoretical analysis showing the convergence of PSL yet, which demands further research efforts.

\vspace{-0.2cm}
\subsection{Asynchronous Split Edge Learning}
\vspace{-0.1cm}
In the current SFL/PSL framework, an edge server updates the model only when accomplishing the training for all clients. However, when an edge device requires much longer training latency or transmission latency due to harsh channel conditions, others have to wait. Asynchronous PSL enables the server to update the server-side model as long as it completes training for one or a given number of clients, thereby boosting resource utilization. However, similar to asynchronous FL, this process potentially hinders model convergence because the stragglers will be under-represented due to less participation in model updates. Consequently, it is crucial to manage "model staleness" in asynchronous SFL and PSL by selecting the appropriate model aggregation frequency, which should adapt to the resource heterogeneity at the wireless edge.

% \subsection{Edge Split Learning with Early Exiting} 
% DNNs usually consist of many interconnected layers with millions of parameters, thus slowing down model inference. To expedite model inference, branches with early exits can be inserted into the hidden layer of the neural network to output prediction results with varied precision. In general, the exits at deeper layers can produce more confident and precise results at the cost of longer latency. The multi-exit DNNs facilitate the acceleration of model inference by providing early inference output with tolerable accuracy. In SEL, a client can first calculate the coarse result based on its client-side sub-model while waiting for the more precise output from the server-side sub-model to correct its prediction if the two results are different. This paradigm, called split inference/computing with early exiting, plays a crucial role in a range of latency-sensitive applications, such as robotic control, as specified in 3GPP technical report~\cite{Release18}. In such a case, the model can be trained in a way that all the (early) classifiers in a model are trained simultaneously by minimizing the summation of all the losses~\cite{han2022splitgp}.

\vspace{-0.2cm}
\subsection{Split Edge Learning with Label Privacy Preservation}
\vspace{-0.1cm}
In conventional SL, labels should be placed on the server side. However, the data label sometimes contains private-sensitive information (i.e., the disease a patient may have), which must be preserved from the edge server. To overcome this, U-shaped split learning has been proposed in \cite{vepakomma2018split}, where both the first and last layers are placed on the client side, allowing the output layers and their respective labels to remain local. \rev{However, this paradigm introduces additional communication costs due to the presence of an extra split point, which necessitates careful selection of two split layers, as well as the effective management of additional data transfer over wireless networks.}

% The corresponding network resource optimization problem, therefore, demands the careful selection of two split layers, as well as the effective management of additional data transfer over wireless networks. However, since this paradigm introduces additional communication costs due to the presence of an extra split point, it is more suitable for application scenarios where label privacy is a major concern.

% \subsection{Multi-view Edge Split Learning} Different devices may collect data corresponding to the same object yet with different statistical properties (e.g., multiple cameras capture images for the same object). Towards this end, multi-view learning has attracted great attention, including the popular multi-view convolutional neural networks based on the fusion of features from multiple devices. By employing over-the-air computation to realize average and max pooling, a multi-view split inference framework has been proposed to improve inference accuracy in \cite{liu2023over}. Apart from inference, it is also crucial to investigate multi-view SL under wireless networks to enhance training performance for multi-view tasks.

\vspace{-0.2cm}
\section{Conclusions\label{conclusion}}
\vspace{-0.1cm}
In the era of 6G, we anticipate that split edge learning can significantly lower the resource demand for on-device model training, allowing for rapid expansion of machine learning across massive IoT devices. This article reviewed the recent advancements in SL and articulated its seamless integration with the 6G edge from both learning and communication perspectives. As a field that remains largely uncharted, a rich set of research opportunities exist, such as the development of more effective and efficient SL frameworks and resource allocation strategies tailored for SL. We hope this work can attract attention from research communities, AI sectors, telecommunication industries, and standardization bodies, ultimately transforming SEL into a tangible reality in the forthcoming 6G era.

\bibliographystyle{IEEEtran}
\bibliography{chen}

\begin{thebibliography}{10}
\providecommand{\url}[1]{#1}
\csname url@samestyle\endcsname
\providecommand{\newblock}{\relax}
\providecommand{\bibinfo}[2]{#2}
\providecommand{\BIBentrySTDinterwordspacing}{\spaceskip=0pt\relax}
\providecommand{\BIBentryALTinterwordstretchfactor}{4}
\providecommand{\BIBentryALTinterwordspacing}{\spaceskip=\fontdimen2\font plus
\BIBentryALTinterwordstretchfactor\fontdimen3\font minus \fontdimen4\font\relax}
\providecommand{\BIBforeignlanguage}[2]{{%
\expandafter\ifx\csname l@#1\endcsname\relax
\typeout{** WARNING: IEEEtran.bst: No hyphenation pattern has been}%
\typeout{** loaded for the language `#1'. Using the pattern for}%
\typeout{** the default language instead.}%
\else
\language=\csname l@#1\endcsname
\fi
#2}}
\providecommand{\BIBdecl}{\relax}
\BIBdecl

\bibitem{Release18}
{3GPP}. ``{{Study on Traffic Characteristics and Performance Requirements for AI/ML Model Transfer in 5GS}}". 3rd Generation Partnership Project (3GPP), Technical Specification (TS) 22.874, 2021, version 18.2.0., Dec. 2021.

\bibitem{huawei20216g}
{Huawei}, \emph{{6G: The Next Horizon: From Connected People and Things to Connected Intelligence}}.\hskip 1em plus 0.5em minus 0.4em\relax Cambridge, U.K.: Cambridge Univ. Press, 2021.

\bibitem{thapa2022splitfed}
C.~Thapa, P.~C.~M. Arachchige, S.~Camtepe, and L.~Sun, ``{Splitfed: When Federated Learning Meets Split Learning},'' in \emph{Proc. AAAI}, Feb. 2022.

\bibitem{shao2020communication}
J.~Shao and J.~Zhang, ``{Communication-computation Trade-off in Resource-constrained Edge Inference},'' \emph{{IEEE} Commun. Mag.}, vol.~58, no.~12, pp. 20--26, Dec. 2020.

\bibitem{mcmahan2017communication}
B.~McMahan, E.~Moore, D.~Ramage, S.~Hampson, and B.~A. y~Arcas, ``{Communication-efficient Learning of Deep Networks From Decentralized Data},'' in \emph{Proc. AISTATS}, Apr. 2017.

\bibitem{vepakomma2018split}
P.~Vepakomma, O.~Gupta, T.~Swedish, and R.~Raskar, ``{Split Learning for Health: Distributed Deep Learning Without Sharing Raw Patient Data},'' \emph{arXiv preprint arXiv:1812.00564}, Dec. 2018.

\bibitem{kim2022bargaining}
M.~Kim, A.~DeRieux, and W.~Saad, ``{A Bargaining Game for Personalized, Energy Efficient Split Learning over Wireless Networks},'' in \emph{Proc. WCNC}, Mar. 2023.

\bibitem{joshi2021splitfed}
P.~Joshi, C.~Thapa, S.~Camtepe, M.~Hasanuzzamana, T.~Scully, and H.~Afli, ``{Splitfed Learning Without Client-side Synchronization: Analyzing Client-side Split Network Portion Size to Overall Performance},'' \emph{arXiv preprint arXiv:2109.09246}, Sep. 2021.

\bibitem{lin2023efficient}
Z.~Lin, G.~Zhu, Y.~Deng, X.~Chen, Y.~Gao, K.~Huang, and Y.~Fang, ``{Efficient Parallel Split Learning over Resource-constrained Wireless Edge Networks},'' \emph{arXiv preprint arXiv:2303.15991}, Mar. 2023.

\bibitem{lin2023pushing}
Z.~Lin, G.~Qu, Q.~Chen, X.~Chen, Z.~Chen, and K.~Huang, ``{Pushing Large Language Models to the 6G Edge: Vision, Challenges, and Opportunities},'' \emph{arXiv preprint arXiv:2309.16739}, Oct. 2023.

\bibitem{hsieh2022c3}
C.-Y. Hsieh, Y.-C. Chuang, and A.-Y. Wu, ``{{C3-SL}: Circular Convolution-Based Batch-Wise Compression for Communication-Efficient Split Learning},'' in \emph{Proc. MLSP}, Aug. 2022.

\bibitem{deng2020model}
L.~Deng, G.~Li, S.~Han, L.~Shi, and Y.~Xie, ``{Model Compression and Hardware Acceleration for Neural Networks: A Comprehensive Survey},'' \emph{Proc IEEE Inst Electr Electron Eng}, vol. 108, no.~4, pp. 485--532, Apr. 2020.

\bibitem{wu2023split}
W.~Wu, M.~Li, K.~Qu, C.~Zhou, X.~Shen, W.~Zhuang, X.~Li, and W.~Shi, ``{Split Learning over Wireless Networks: Parallel Design and Resource Management},'' \emph{{IEEE} J. Sel. Areas Commun.}, vol.~41, no.~4, pp. 1051--1066, Apr. 2023.

\bibitem{chen2022federated}
X.~Chen, G.~Zhu, Y.~Deng, and Y.~Fang, ``{Federated Learning over Multihop Wireless Networks With In-Network Aggregation},'' \emph{IEEE Trans. Wirel. Commun.}, vol.~21, no.~6, pp. 4622--4634, Apr. 2022.

\bibitem{9562523}
S.~Wang, X.~Zhang, H.~Uchiyama, and H.~Matsuda, ``{HiveMind: Towards Cellular Native Machine Learning Model Splitting},'' \emph{{IEEE} J. Sel. Areas Commun.}, vol.~40, no.~2, pp. 626--640, Feb. 2022.

\end{thebibliography}

\begin{IEEEbiographynophoto}{Zheng Lin} is currently pursuing his Ph.D. degree at the University of Hong Kong. His research interests include edge intelligence and distributed machine learning.
\end{IEEEbiographynophoto}

\begin{IEEEbiographynophoto}{Guanqiao Qu} is currently pursuing his Ph.D. degree at the University of Hong Kong. His research interests include edge intelligence and federated learning.
\end{IEEEbiographynophoto}

\begin{IEEEbiographynophoto}{Xianhao Chen} is an assistant professor at the Department of Electrical and Electronic Engineering, The University of Hong Kong. His research interests include wireless networking, edge intelligence, and distributed learning.
\end{IEEEbiographynophoto}

\begin{IEEEbiographynophoto}{Kaibin Huang} [Fellow, IEEE] is a Professor at the Department of Electrical and Electronic Engineering, The University of Hong Kong. His research interests include mobile edge computing, edge AI, and 6G systems.
\end{IEEEbiographynophoto}

%\vfill

\end{document}